%% file: main.tex
\documentclass[10pt,twocolumn,letterpaper]{article}

\usepackage[letterpaper,margin=0.75in,columnsep=0.28in]{geometry}
\usepackage[T1]{fontenc}
\usepackage{times}
\usepackage{amsmath,amssymb}
\usepackage{graphicx}
\usepackage{array,booktabs,colortbl}
\usepackage[font=small,labelfont=bf,labelsep=period]{caption}
\usepackage[numbers,sort&compress]{natbib}
\usepackage{cuted}
\usepackage[hidelinks]{hyperref}

\newcommand{\method}{LightFuse}
\newcommand{\vect}[1]{\mathbf{#1}}
\newcommand{\Real}{\mathbb{R}}

\newcommand{\suppsec}[1]{Supp.~Sec.~\ref*{S-#1}}
\newcommand{\suppsecs}[2]{Supp.~Secs.~\ref*{S-#1} and~\ref*{S-#2}}
\title{\method{}: Relightable Interactive Gaussian Scene Reconstruction via Multi-Scan Fusion and 2D Gaussian Ray Tracing}
\author{Haonan Zhou\textsuperscript{1},
    Gaoxiang Linghu\textsuperscript{1},
    Youlin Jia\textsuperscript{2},
    Chuanyu Cheng\textsuperscript{1}\\
    Hongyu Cui\textsuperscript{1},
    Kewei Wei\textsuperscript{1},
    Bruce X.B. Yu\textsuperscript{1},
    Gaoang Wang\textsuperscript{1}\thanks{Corresponding author.}\\[0.5em]
    \small \textsuperscript{1}\,Zhejiang University\\[-0.1em]
    \small \textsuperscript{2}\,Sichuan University
}
\date{}

\hypersetup{
    pdftitle={LightFuse: Relightable Interactive Gaussian Scene Reconstruction via Multi-Scan Fusion and 2D Gaussian Ray Tracing},
    pdfauthor={Haonan Zhou, Gaoxiang Linghu, Youlin Jia, Chuanyu Cheng, Hongyu Cui, Kewei Wei, Bruce X.B. Yu, Gaoang Wang}
}

\begin{document}
\maketitle

\begin{strip}
    \centering
    \includegraphics[width=\linewidth, trim=0 0 0 0, clip]{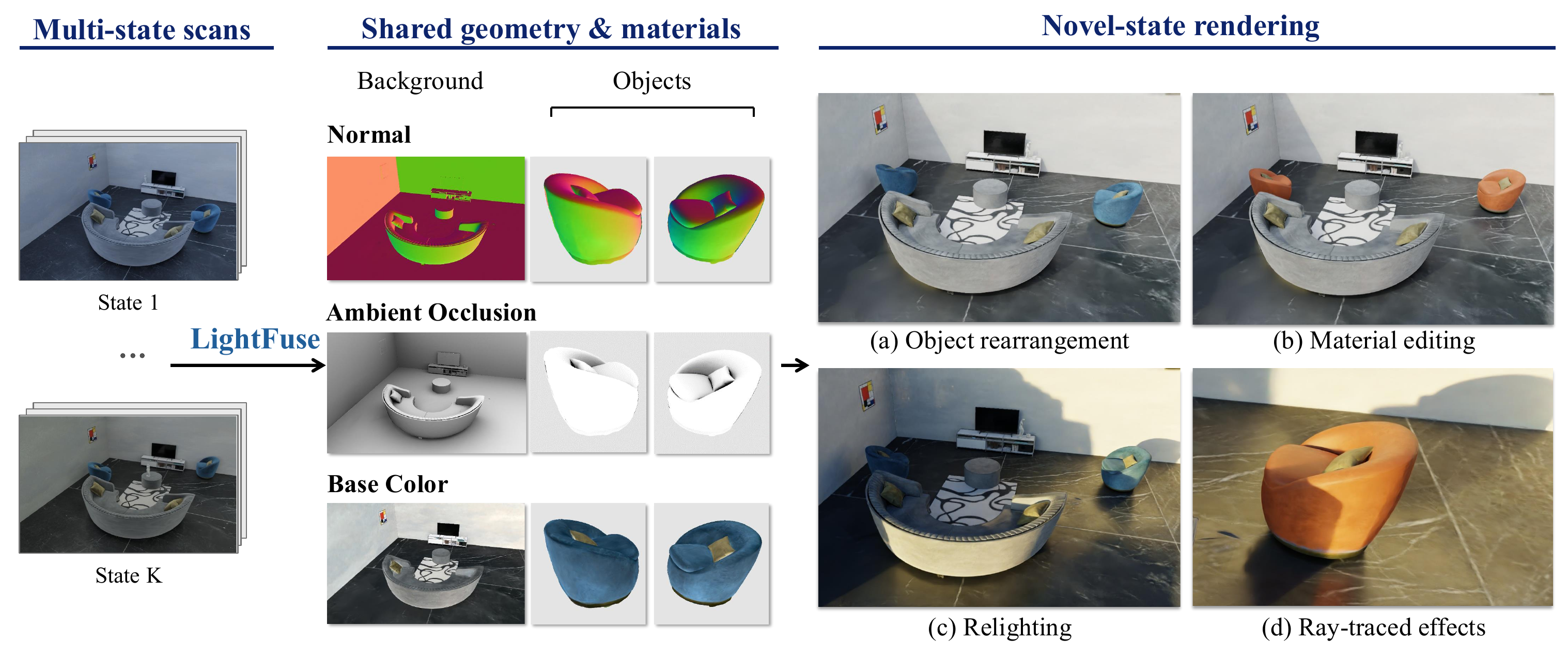}
    \captionsetup{hypcap=false}
    \captionof{figure}{\method{} reconstructs explicit geometry and materials for (a) rearrangement, (b) material editing, (c) relighting, and (d) ray-traced effects.}
    \captionsetup{hypcap=true}
    \label{fig:teaser}
\end{strip}

\input{sections/abstract}
\input{sections/introduction}
\input{sections/related_work}
\input{sections/method}
\input{sections/experiments}
\input{sections/discussion_limitations}
\input{sections/conclusion}

\clearpage
\input{sections/llm_usage}
{
    \small
    \bibliographystyle{ieeenat_fullname}
    \bibliography{lightfuse_refs}
}
\end{document}

%% file: sections/abstract.tex
\begin{abstract}
Relightable interactive scene reconstruction aims to build an editable 3D model from scans of different object arrangements, so that novel layouts can be rendered under novel illumination.
However, Gaussian-splatting-based methods bake illumination into appearance and thus cannot jointly reconstruct scans captured under different lighting, while existing multi-illumination inverse rendering methods struggle to handle scenes whose geometry changes across captures.
We present \textbf{LightFuse}, a 2D Gaussian inverse rendering framework that bridges this gap by extending interactive scene reconstruction with explicit material-illumination decomposition and physically based relighting.
LightFuse fuses observations across states to reconstruct a shared background and movable objects. 
Since geometry optimized by splatting alone is ill-suited for ray-tracing-based inverse rendering, it then performs ray-tracing-oriented geometry refinement to obtain more complete and consistent surfaces.
Because rearrangement leaves materials unchanged, we jointly optimize a shared set of materials across all scans under varying illumination, with differentiable one-bounce ray tracing disentangling them from the per-scan environment lighting.
Experiments on novel-state relighting of synthetic scenes and novel-state synthesis of captured real scenes show that LightFuse outperforms interactive reconstruction and adapted inverse rendering baselines, improving novel-state relighting PSNR by 6.01 dB over the strongest baseline and achieving the highest PSNR on captured scenes.
\mbox{Project page: \urlstyle{same}\url{https://zhn202.github.io/LightFuse/}.}
\end{abstract}

%% file: sections/introduction.tex
\section{Introduction}

\begin{figure}[t]
    \centering
    \includegraphics[width=\columnwidth]{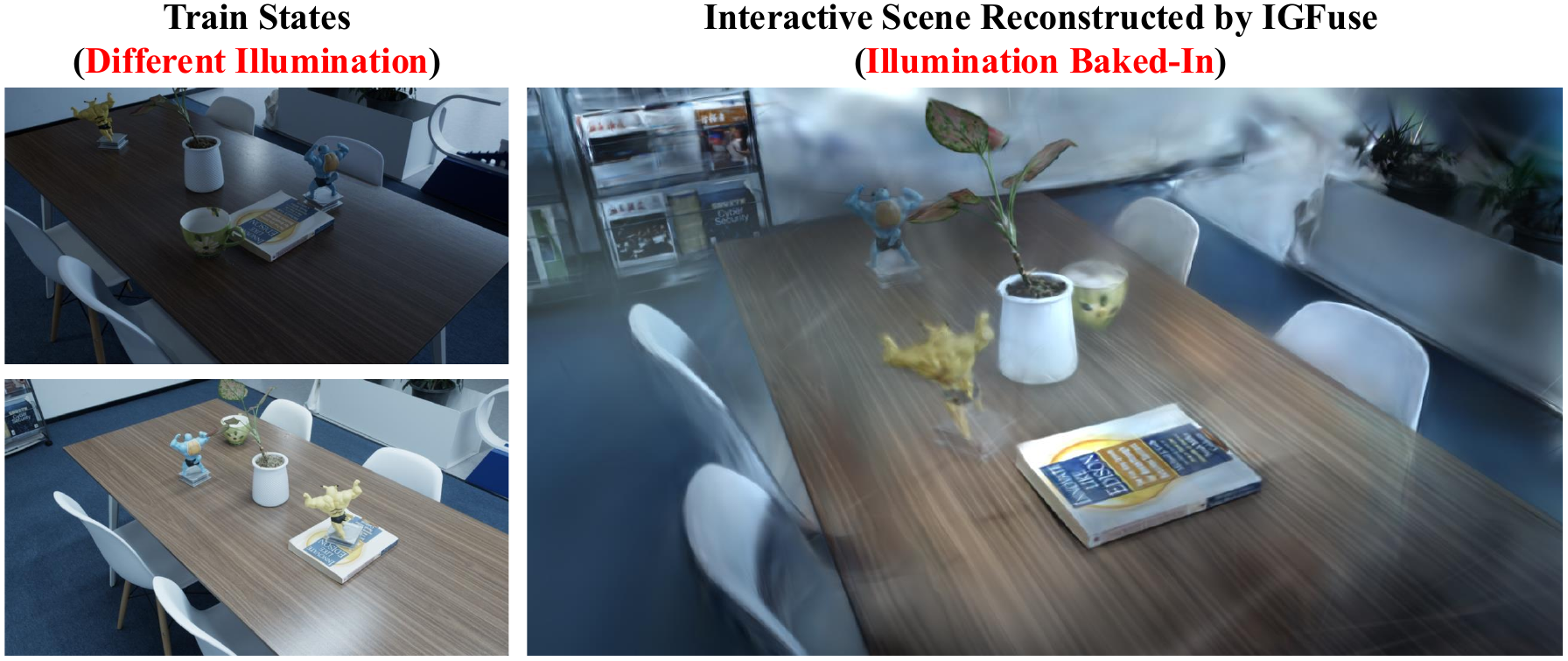}
    \caption{Photometric inconsistency in interactive scene reconstruction. IGFuse~\citep{hu2025igfuse} fuses scans captured under different illumination into a shared representation, so the reconstructed scene carries baked-in, inconsistent appearance that does not match any single state.}
    \label{fig:igfuse_demo}
\end{figure}

Interactive scene reconstruction builds an editable 3D model from multiple scans in which objects are rearranged between captures \citep{hu2025igfuse}.
The recovered objects can be moved to create new scene layouts, which is useful for embodied simulation \citep{szot2021habitat}, robotic learning \citep{qureshi2025splatsim,tian2023multiobject}, virtual staging \citep{ji2024virtualhome}, and mixed reality \citep{li2022physically}.
These applications call for more than rearrangement: the edited scene should remain physically consistent, and ideally support relighting under novel illumination and editing of object materials \citep{li2022physically,gao2024relightable}.

Existing interactive-scene methods based on Gaussian splatting \citep{kerbl2023three,huang2024twodgs} fuse scans with different object arrangements to separate movable objects from the background and complete unseen geometry \citep{hu2025igfuse,hu2025recurgs,kim2026ltgs}.
However, they optimize appearance directly from the images, entangling material and illumination in the same Gaussians, which rules out relighting and material editing and causes two failures.
First, appearance that varies with arrangement and illumination across states is baked into a shared representation, producing photometric inconsistencies (Figure~\ref{fig:igfuse_demo}).
Second, in novel layouts, baked-in shadows, highlights, and indirect illumination stay fixed to the objects instead of adapting, resulting in physically inconsistent lighting.

Inverse rendering with neural and Gaussian representations decomposes appearance into material and illumination \citep{zhang2023neilfpp,mai2023neuralmicrofacet,liu2023nero,liang2024gsir,gu2025irgs,liang2025gusir}, and can exploit multiple illuminations to reduce their ambiguity \citep{jin2023tensoir,li2025recap,bai2025gare,kaleta2025lumigauss}, yet it assumes a static scene whose geometry is shared by all captures.
In interactive scenes, illumination and object layout change together across scans, so the same surface is observed under different lighting at different positions, and existing methods cannot associate these observations.

Our key insight is that the background and movable objects retain their intrinsic geometry and materials despite changes in layout and illumination. Sharing geometry and materials provides a scene representation across states. This representation supports scene completion from complementary observations and constrains material–illumination decomposition through appearance variations.

We propose \textbf{\method{}}, which realizes this insight in three stages.
It decomposes the reconstruction of a reference state into a shared background and movable objects, which can then be placed in any layout.
We complete the background and objects under multi-state depth and normal supervision.
Since geometry optimized by splatting alone is ill-suited for ray-tracing-based inverse rendering, we perform ray-tracing-oriented refinement to obtain more consistent surfaces.
With the refined geometry fixed, we jointly optimize shared material parameters and per-state environment maps via differentiable one-bounce path tracing, progressively increasing samples per pixel during training for efficiency.
The resulting scene supports object rearrangement, relighting, and material editing (Figure~\ref{fig:teaser}).

Our contributions are:
\begin{itemize}
    \item We present LightFuse, to our knowledge the first framework for relightable interactive Gaussian reconstruction, enabling physically consistent novel-state synthesis.
    \item We propose a ray-tracing-oriented geometry optimization scheme that mitigates spurious occlusions and light leakage during ray tracing of Gaussian scenes.
    \item Our multi-state inverse rendering achieves more faithful material--illumination decomposition. In novel-state relighting experiments, \method{} achieves higher rendering quality than baseline methods, producing more realistic results under novel layouts and illumination.
\end{itemize}

%% file: sections/related_work.tex
\begin{figure*}[t]
    \centering
    \includegraphics[width=\textwidth]{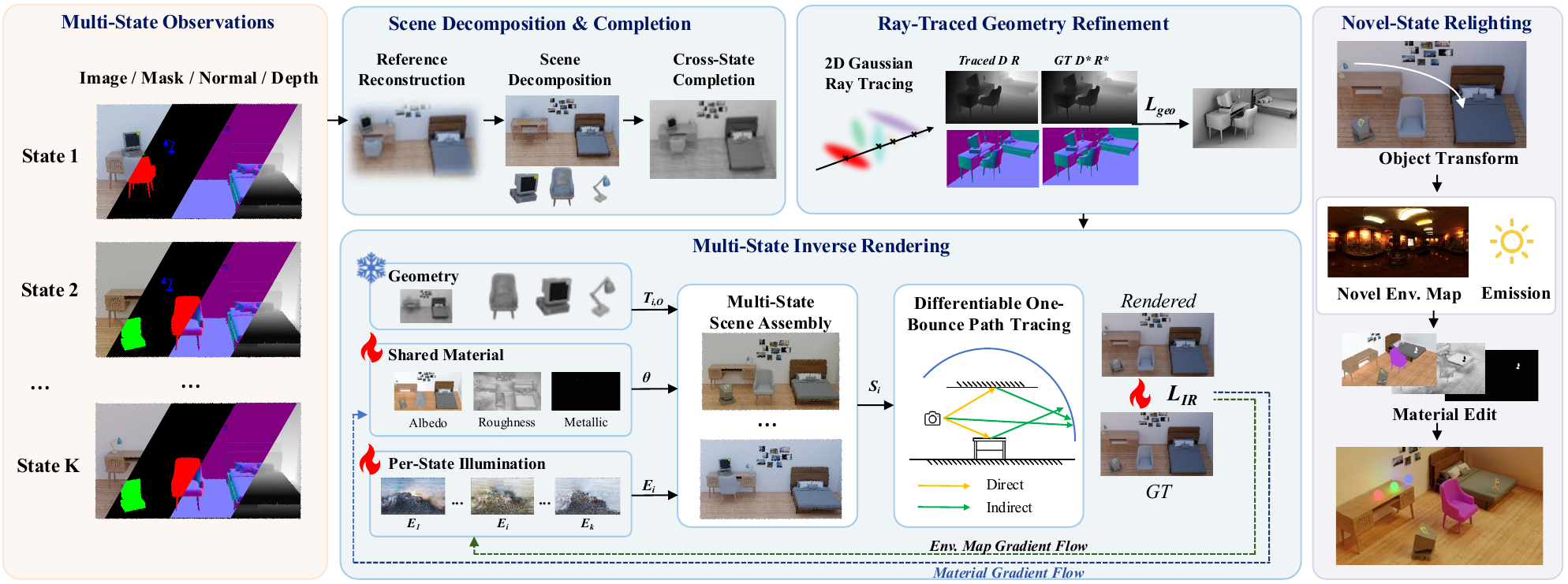}
    \caption{Pipeline of \method{}. First, multi-state scans are fused into an occlusion-complete background and movable objects. The shared 2D Gaussian surfaces are then refined for ray tracing through cross-state supervision and geometric regularization, after which staged inverse rendering uses differentiable one-bounce path tracing to separate shared material from state-specific environment illumination. The resulting model supports object rearrangement, material editing, and relighting, with visibility and light transport recomputed.}
    \label{fig:pipeline}
\end{figure*}

\section{Related Work}

\subsection{Interactive Gaussian Scenes}

The explicit primitives of Gaussian Splatting support object selection, appearance editing, and geometric manipulation \citep{kerbl2023three,ye2024gaussiangrouping,wang2025decoupledgaussian}.
GaussianEditor propagates localized edits through a reconstructed field, while GaussianCut extracts object-level subsets through interactive graph-cut segmentation \citep{chen2024gaussianeditor,jain2024gaussiancut}.
These methods manipulate existing representations but do not fuse scene rearrangements.

Multi-scan methods use observations across states to decompose objects and background.
IGFuse fuses captures to complete object and background regions occluded in any one arrangement \citep{hu2025igfuse}.
RecurGS incrementally integrates states through object-motion alignment and visibility-aware fusion, while LTGS maintains objects as reusable Gaussian templates from sparse updates \citep{hu2025recurgs,kim2026ltgs}.
They recover manipulable geometry but not shared material and state-specific illumination, and cannot consistently update shadows and inter-reflections after recomposition.
\method{} extends discrete-state fusion to transport-aware recomposition, reevaluating visibility and light transport after edits.

\subsection{Inverse Rendering and Relighting}

Inverse rendering decomposes appearance into geometry, reflectance, and illumination for relighting and material editing \citep{zhang2021physg,yao2022neilf}.
NeRFactor and TensoIR recover these properties with neural fields \citep{zhang2021nerfactor,jin2023tensoir}, whereas Gaussian-based methods encode material and illumination attributes on explicit primitives and optimize them through physically based rendering \citep{gao2024relightable,liang2024gsir,chen2024gigs,zheng2026ssdgs}.
Based on the surface representation of 2D Gaussian Splatting \citep{huang2024twodgs}, IRGS traces Gaussian surfaces to evaluate visibility and inter-reflection \citep{gu2025irgs}.
EAG-PT and PTIR-GS incorporate Gaussian path tracing for reconstruction, inverse rendering, and editing \citep{yang2026eagpt,zhu2026ptirgs}.
These methods assume fixed geometry rather than fuse discrete rearrangements into occlusion-complete objects.

Cross-condition observations provide complementary constraints on material--illumination ambiguity.
ReCap shares Gaussian material across environments while jointly fitting their illumination \citep{li2025recap}.
GauUpdate aligns source and target Gaussian fields to recover shared material under distinct lighting for consistent object insertion \citep{ren2025gauupdate}.
Other methods explore temporal dynamics \citep{fan2025spectromotion,kaleta2026lumimotion} or object--scene composition settings \citep{gao2026comgs,perel2026tron}.
Unlike these prior approaches, \method{} fuses discrete rearranged states and retraces light transport after recomposition or relighting.

%% file: sections/method.tex
\section{Method}

\method{} reconstructs an editable and relightable 2D Gaussian scene from multi-state scans, as illustrated in Figure~\ref{fig:pipeline}.
It first fuses observations across states into a shared background and movable objects, then refines their geometry for ray tracing, and finally separates shared materials from state-specific illumination through staged inverse rendering.
The resulting 3D model allows for object rearrangement, material editing, and relighting.

\subsection{Scene Completion with 2D Gaussian Splatting}
\label{sec:interactive-reconstruction}

Given multi-view observations $\{I_{k,v}\}$ from $K$ scans of the same scene, where $k$ and $v$ index the scene state and view, objects are rearranged and illumination may vary across states.
The same surface can therefore appear differently in different states, breaking the cross-state appearance consistency on which prior interactive reconstruction relies \citep{hu2025igfuse}.
Since rigid objects preserve their geometry under rearrangement, we reconstruct appearance from a single reference state and use the remaining states only as geometric evidence for completing a shared scene.

\paragraph{Scene representation.}
We represent the scene as a shared background $\mathcal B$ and $M$ movable objects $\{\mathcal O_o\}_{o=1}^{M}$, each composed of 2D Gaussian surfels \citep{huang2024twodgs}.
The $i$-th surfel is
\begin{equation}
    \mathcal{G}_i =
    \left(
    \vect{\mu}_i,\vect{q}_i,\vect{s}_i,\alpha_i,
    \vect{a}_i,m_i,r_i,\vect{f}_i
    \right).
\label{eq:gaussian-primitive}
\end{equation}
Here, $\vect{\mu}_i\in\Real^3$, $\vect{q}_i\in\mathbb{S}^3$, $\vect{s}_i\in\Real_+^2$, and $\alpha_i$ denote its center, orientation, in-plane scale, and opacity.
The base color $\vect{a}_i\in[0,1]^3$, metallic $m_i\in[0,1]$, and roughness $r_i\in[0,1]$ parameterize physically based shading, whereas $\vect f_i$ is an auxiliary splatting appearance used only before inverse rendering.
Surfel geometry and material are shared across all states.
Each state $k$ differs only in its object poses and its illumination, modeled as an environment map $E_k$:
\begin{equation}
\mathcal S_k
=
\Bigl(
\mathcal B
\cup
\bigcup_{o=1}^{M}
\mathcal T_{k,o}(\mathcal O_o),\;
E_k
\Bigr),
\qquad k=1,\ldots,K .
\label{eq:state-assembly}
\end{equation}
The rigid transform $\mathcal T_{k,o}=(\vect R,\vect t)$ maps object $o$ from its canonical frame to state $k$, updating each surfel as $\vect\mu_i\mapsto\vect R\vect\mu_i+\vect t$ and $\vect q_i\mapsto\vect q(\vect R)\otimes\vect q_i$.
All other attributes remain unchanged, so a surface observed in any state constrains the same geometry and material in every state, whereas $E_k$ absorbs state-specific illumination.
The geometry is completed and refined in Sec.~\ref{sec:interactive-reconstruction}--\ref{sec:ray-tracing-geometry}, and the material and $\{E_k\}$ are recovered in Sec.~\ref{sec:inverse-rendering}.

\paragraph{Cross-state completion.}
We reconstruct the reference state with RGB and geometric supervision and, following IGFuse \citep{hu2025igfuse}, decompose it into $\mathcal B$ and $\{\mathcal O_o\}$ using SAM~3 masks \citep{carion2025sam3segmentconcepts} and GaussianCut \citep{jain2024gaussiancut}, taking each object's pose in this state as its canonical frame and estimating $\mathcal T_{k,o}$ for the other states.
Surfaces hidden in the reference state, such as the background beneath objects and their contact faces, are exposed by rearrangements in other states, from which we recover them using depth and normals only.
Treating $\mathcal B$ as a component with identity transform, we render each component in every view of every state and mark pixels within its segmentation mask where the rendered depth $\hat D$ lies behind the observed depth $D^*$:
\begin{equation}
\mathcal H=\left\{p \mid \hat D(p)-D^{*}(p)>\tau_D\right\}.
\label{eq:completion-mask}
\end{equation}
These pixels are back-projected with their observed depth and normal into new surfels, mapped to the canonical frame via $\mathcal T_{k,o}^{-1}$, and assigned the observed color as $\vect f_i$, which is discarded before inverse rendering.
The completed scene is then optimized over all states with $\mathcal L_{\mathrm{geo}}$ (Sec.~\ref{sec:training-objective}); registration and completion details are in \suppsec{sec:supp-reconstruction}.

\subsection{Ray-Tracing-Oriented Geometry Optimization}
\label{sec:ray-tracing-geometry}

The reconstructed geometry ultimately serves ray-traced inverse rendering, in which shadow and secondary rays query surfaces along directions never observed by the cameras.
We therefore first describe the 2D Gaussian tracer shared by refinement and inverse rendering, then regularize the splatting-based stages against defects that the tracer exposes but splatting does not penalize, and finally refine the completed geometry with the tracer itself.

\begin{figure}[t]
    \centering
    \includegraphics[width=\columnwidth]{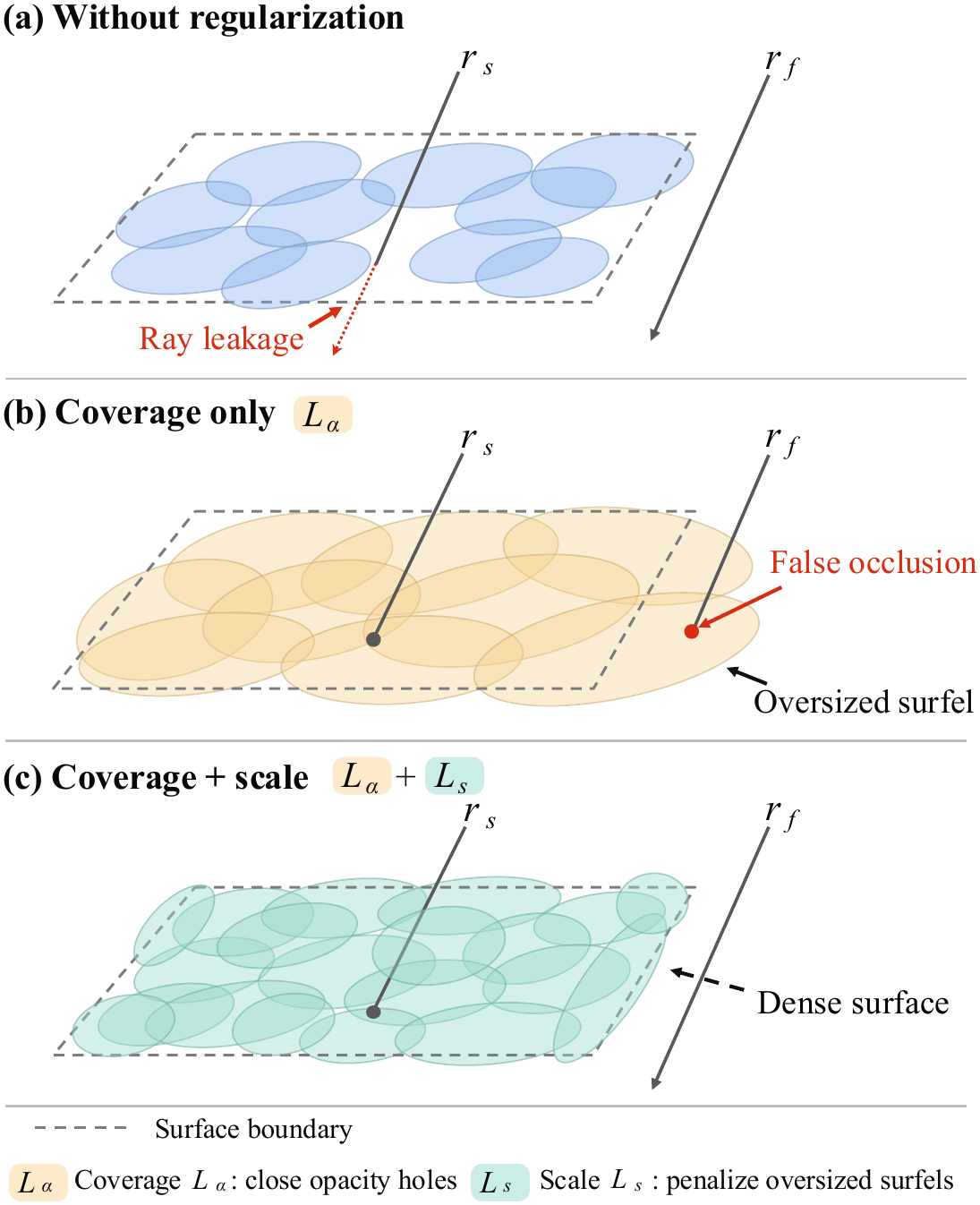}
    \caption{Effect of the regularizers on a surfel patch, with a shadow ray $r_s$ and a free-space ray $r_f$. (a) Opacity holes let $r_s$ leak. (b) $\mathcal L_\alpha$ closes holes with oversized surfels, falsely occluding $r_f$. (c) $\mathcal L_s$ bounds surfel size, blocking $r_s$ without occluding $r_f$.}
    \label{fig:reg}
\end{figure}

\paragraph{2D Gaussian ray tracing.}
Following prior Gaussian ray-tracing methods \citep{gu2025irgs,yang2026eagpt,xie2024envgs,moenne2024gaussianraytracing}, each surfel is treated as an oriented tracing primitive with finite support.
For a ray $\vect o+t\vect d$, candidate surfel planes are intersected, and each intersection is projected onto the corresponding tangent axes to obtain its normalized local coordinate $\boldsymbol{\xi}_j\in\Real^2$.
The effective opacity is
\begin{equation}
\widetilde\alpha_j
=
\alpha_j
\exp\!\left(-\frac{1}{2}\|\boldsymbol{\xi}_j\|_2^2\right),
\label{eq:gaussian-response}
\end{equation}
capped at $0.99$ for numerical stability; intersections outside the finite support or below the opacity threshold are discarded.
The remaining intersections are sorted by plane-intersection depth and composited front to back:
\begin{equation}
    T_j=\prod_{\ell<j}(1-\widetilde{\alpha}_\ell),\quad
    w_j=T_j\widetilde{\alpha}_j,\quad
    \bar{\vect z}=
    \frac{\sum_j w_j\vect z_j}{\sum_j w_j+\epsilon}.
    \label{eq:surface-record}
\end{equation}
For the $j$-th intersection with depth $d_j$, setting $\vect z_j$ to its hit position $\vect x_j=\vect o+d_j\vect d$, $d_j$, $\vect n_j$, or $(\vect a_j,m_j,r_j)$ yields the composited position, depth, normal, or material, and $\sum_j w_j$ gives the accumulated opacity.
The composited normal is renormalized before shading, and the shared weights keep geometry and material aligned across overlapping surfels.

\begin{figure*}[!t]
    \centering
    \includegraphics[width=\textwidth]{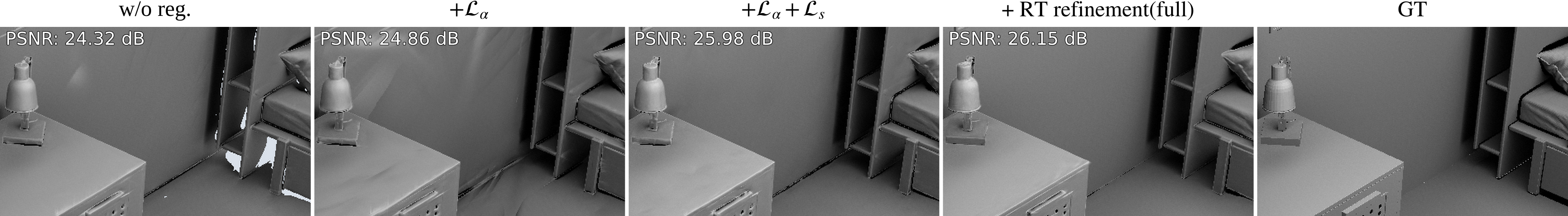}
    \caption{Ambient occlusion of the reconstructed geometry on one held-out view, computed with our 2D Gaussian ray tracer.
    Under splatting-based reconstruction, opacity holes leak occlusion without regularization; $\mathcal L_\alpha$ closes these holes but inflates surfels, producing spurious occlusion that $\mathcal L_s$ suppresses.
    Ray-traced geometric refinement further corrects surfaces that appear accurate under splatting but are inconsistent under ray tracing.
    }
    \label{fig:ao_geometry}
\end{figure*}

\paragraph{Coverage and scale regularization.}
Splatting only needs to reproduce the training views and therefore does not penalize two defects that matter to the tracer.
First, surfels are discrete and semi-transparent rather than closed surfaces, so opacity holes leave the transmittance $T_j$ high and let shadow and secondary rays leak through.
We encourage opaque coverage using
\begin{equation}
\mathcal{L}_{\alpha}
=
\frac{1}{HW}
\left\|
\hat{\vect A}-\vect 1
\right\|_1,
\label{eq:alpha-coverage}
\end{equation}
where $\hat{\vect A}$ is the accumulated opacity map.
Second, $\mathcal L_\alpha$ alone can be satisfied by enlarging surfels, especially after densification stops, and oversized surfels extend into free space and create false occlusions.
We constrain this growth using
\begin{equation}
\mathcal{L}_s
=
\frac{1}{N}
\sum_{i=1}^{N}
\left[
\max\left(
0,\frac{\|\vect{s}_i\|_{\infty}}{s_{\max}}-1
\right)
\right]^2 ,
\label{eq:scale-regularization}
\end{equation}
where $N$ is the number of surfels, $\|\vect{s}_i\|_{\infty}$ selects the largest in-plane scale of the $i$-th surfel, and $s_{\max}$ is the maximum permitted extent.
The two terms are complementary (Figure~\ref{fig:reg}): $\mathcal L_\alpha$ closes holes, while $\mathcal L_s$ prevents closing them with oversized, geometrically imprecise surfels.
Since shadow and secondary rays cannot be supervised directly, these terms act as proxies that constrain how the geometry responds to unobserved ray directions.

\paragraph{Ray-traced geometric refinement.}
Unlike splatting, which composites surfels in an order determined by their center depths, the tracer orders contributions by per-ray intersection depth and applies finite support and opacity thresholds.
Geometry that reproduces the training views under splatting may therefore yield different depth, normals, and visibility under ray tracing (Fig.~\ref{fig:ao_geometry}).
Since splatting remains more efficient for densification and appearance reconstruction, we close this gap with a short refinement before inverse rendering: every assembled state $\mathcal S_k$ is traced from its observed views, and the shared surfels are updated without RGB supervision by $\mathcal L_\alpha$, $\mathcal L_s$, and standard depth and normal losses \citep{huang2024twodgs}:
\begin{equation}
\mathcal L_D=\frac{\|C\odot(\hat D-D^*)\|_1}{\|C\odot D^*\|_1+\epsilon},\qquad
\mathcal L_N(\vect N)=1-\langle\vect N,\vect N^*\rangle,
\label{eq:geometry-losses}
\end{equation}
where $C$ is a depth-confidence map, $\vect N^*$ is the observed normal map, and $\langle\cdot,\cdot\rangle$ averages the per-pixel inner product.
$\mathcal L_N$ is applied to both the rendered normals $\hat{\vect N}$ and the normals $\hat{\vect N}^D$ derived from rendered depth.
The same losses are evaluated with splatting during reconstruction and completion, whereas here all quantities come from the tracer (Eq.~\eqref{eq:surface-record}), so the geometry used for inverse rendering is supervised under exactly the intersection and compositing rules with which it is later traced.

\subsection{Multi-State Inverse Rendering}
\label{sec:inverse-rendering}

With the refined geometry fixed, we recover the shared materials $(\vect a,m,r)$ and per-state environment maps $\{E_k\}$ from all states.
Since rearranging objects moves their shadows and changes inter-reflections with the background, visibility and indirect illumination vary across states and cannot be baked into the surfels or taken from the reference-state appearance.
We trace each assembled state $\mathcal S_k$, so that state-specific appearance changes must be explained by $E_k$ and the shared scene alone, which lets observations across states constrain the material--illumination decomposition.

\paragraph{One-bounce path tracing.}
At a traced surface point $\vect x$ (Eq.~\eqref{eq:surface-record}), we evaluate the rendering equation \citep{kajiya1986rendering}
\begin{equation}
L_o(\vect x,\vect\omega_o)=\int_{\Omega^+} f_r(\vect\omega_i,\vect\omega_o)\,L_i(\vect x,\vect\omega_i)\,(\vect n^\top\vect\omega_i)\,d\vect\omega_i
\label{eq:rendering-equation}
\end{equation}
with a metallic--roughness BRDF $f_r$ combining Lambertian diffuse and GGX specular terms \citep{munkberg2022extracting,walter2007microfacet}.
Let $\widehat L_{\mathrm{dir}}(\vect x)$ denote the direct environment estimate at $\vect x$, combining next-event estimation (NEE) and BSDF sampling of $E_k$ with multiple importance sampling (MIS) \citep{veach1995optimally}.
As a trade-off between accuracy and cost, we limit paths to one secondary interaction:
\begin{equation}
\widehat L(\vect x)=\widehat L_{\mathrm{dir}}(\vect x)+\frac{1}{S}\sum_{s=1}^{S} T_s\,\widehat L_{\mathrm{dir}}(\vect y_s),
\label{eq:path-estimator}
\end{equation}
where $\vect y_s$ is the secondary hit of the $s$-th BSDF-sampled ray and $T_s$ is its BRDF--cosine--density throughput; rays that escape contribute only through $\widehat L_{\mathrm{dir}}(\vect x)$.

\paragraph{Optimization.}
We supervise the estimator in Eq.~\eqref{eq:path-estimator} against the observed images in scene-linear HDR space,
\begin{equation}
\mathcal L_{\mathrm{IR}}=\mathcal L_{\mathrm{HDR}}+\lambda_E\sum_{k}\mathcal L_{\mathrm{TV}}(E_k),
\label{eq:training-objectives}
\end{equation}
where $\mathcal L_{\mathrm{HDR}}$ is an $\ell_1$ photometric loss that preserves the dynamic range of highlights and shadows, and the total-variation term $\mathcal L_{\mathrm{TV}}$ suppresses high-frequency noise in each $E_k$.
Following path replay backpropagation \citep{vicini2021pathreplay,zhu2026ptirgs}, we replay the forward samples in the backward pass, with gradients flowing through the primary BRDF and $E_k$ while sampling decisions and secondary records are held fixed.
Since tracing dominates the cost, we increase the samples per pixel from 8 to 16 to 32 during training to reduce early-stage sampling cost; Sec.~\ref{sec:ablations} compares this schedule with fixed-rate training.

\subsection{Training and Editing}
\label{sec:training-inference}

\paragraph{Training.}
\label{sec:training-objective}
Training proceeds through reference reconstruction, cross-state completion, ray-tracing-oriented geometry refinement, and inverse rendering.
The geometry and reconstruction objectives are
\begin{equation}
\begin{aligned}
\mathcal L_{\mathrm{geo}}
={}&\lambda_\alpha\mathcal L_\alpha
+\lambda_s\mathcal L_s
+\lambda_D\mathcal L_D\\
&+\lambda_N\mathcal L_N(\hat{\vect N})
+\lambda_{DN}\mathcal L_N(\hat{\vect N}^D),\\
\mathcal L_{\mathrm{rec}}
={}&\mathcal L_{\mathrm{RGB}}+\mathcal L_{\mathrm{geo}},
\end{aligned}
\label{eq:geometry-objective}
\end{equation}
where $\mathcal L_{\mathrm{RGB}}$ combines $\ell_1$ and DSSIM losses.
$\mathcal L_{\mathrm{rec}}$ is evaluated with splatting during reconstruction and completion (without $\mathcal L_{\mathrm{RGB}}$ for completion), $\mathcal L_{\mathrm{geo}}$ with the tracer during refinement, and $\mathcal L_{\mathrm{IR}}$ (Eq.~\eqref{eq:training-objectives}) during inverse rendering.
Stage-specific weights and optimization settings are provided in \suppsec{sec:supp-training}.

\paragraph{Editing and relighting.}
\label{sec:editing-relighting}
Our componentized scene supports object rearrangement under target rigid transforms $\{\mathcal T_o^\star\}_{o=1}^{M}$ and material edits such as base-color replacement and metallic or roughness adjustment.
The edited scene is re-traced with Eq.~\eqref{eq:path-estimator} under a recovered $E_k$ or a user-provided HDR environment, so shadows and inter-reflections follow the new layout and illumination.

%% file: sections/experiments.tex
\section{Experiments}

\subsection{Dataset}
\label{sec:datasets}

For \emph{novel-state relighting}, we use four Blender scenes---Bedroom, Playroom, Livingroom, and Kitchen---built from CL-Splats assets and BlenderKit \citep{blender2023,ackermann2025clsplats,blenderkit2023}.
Each has three training states with different layouts and lighting, and one held-out layout under eleven unseen environment maps.
For \emph{novel-state synthesis}, we use two captured real scenes, Captured 1 and Captured 2, whose training states differ in both layout and lighting; each test state changes only object poses and retains the lighting of one training state.
Since real illumination cannot be measured as ground truth, real scenes are evaluated for synthesis rather than relighting.
Each inverse renderer renders the test state with its own recovered environment map of the corresponding training state, without fitting any test image.
We report PSNR and SSIM.
Splits are detailed in \suppsec{sec:supp-splits}, lighting correspondence in \suppsec{sec:supp-eval-tasks}, and metrics in \suppsec{sec:supp-metrics}.

\subsection{Experimental Setup}
\label{sec:experimental-setup}

\paragraph{Implementation details.}
We conduct inverse rendering for 15,000 iterations, sampling 16,384 pixels from one image per iteration.
Each observed state contains a trainable $256\times128$ RGB HDR environment map.
The sampling rate increases from 8 samples per pixel (spp) to 16 spp at iteration 7,500 and to 32 spp at iteration 12,000.
Each path uses one primary NEE sample of the environment and one BSDF continuation, combined with MIS, followed by at most one secondary hit with four environment samples.
Unless otherwise stated, evaluation uses 32 spp.
Remaining preprocessing and optimization details are in \suppsecs{sec:supp-datasets}{sec:supp-training}, and tracing details in \suppsec{sec:supp-ray-geometry}.
For synthetic datasets, reference depth and normal maps are rendered in Blender.
For real-world datasets, reference depth and normal maps are estimated using Depth Anything 3 \citep{lin2025depthanything3} and StableNormal \citep{ye2024stablenormal}, respectively.
For evaluation, we denoise surface radiance using the OptiX HDR denoiser with base color and normal guides.
All experiments run on a single NVIDIA GeForce RTX 4090 GPU.

\paragraph{Baselines.}
We compare against IGFuse \citep{hu2025igfuse}, ReCap \citep{li2025recap}, and IRGS \citep{gu2025irgs}.
IGFuse cannot accept a target illumination, whereas ReCap and IRGS do not support multi-state, multi-illumination inputs.
We therefore adapt ReCap and IRGS to our setting: they receive the same object/background decomposition, rigid transforms $\mathcal T_{k,o}$, per-state environment maps $E_k$, and cross-state shared materials as \method{}, while keeping their own geometry and rendering models.
Adaptation details are provided in \suppsecs{sec:supp-recap-adaptation}{sec:supp-irgs-adaptation}, and the geometry-transfer variant in \suppsec{sec:supp-geometry-transfer}.

\subsection{Novel-State Relighting}
\label{sec:synthetic-results}

\begin{figure*}[t]
    \centering
    \includegraphics[width=\textwidth]{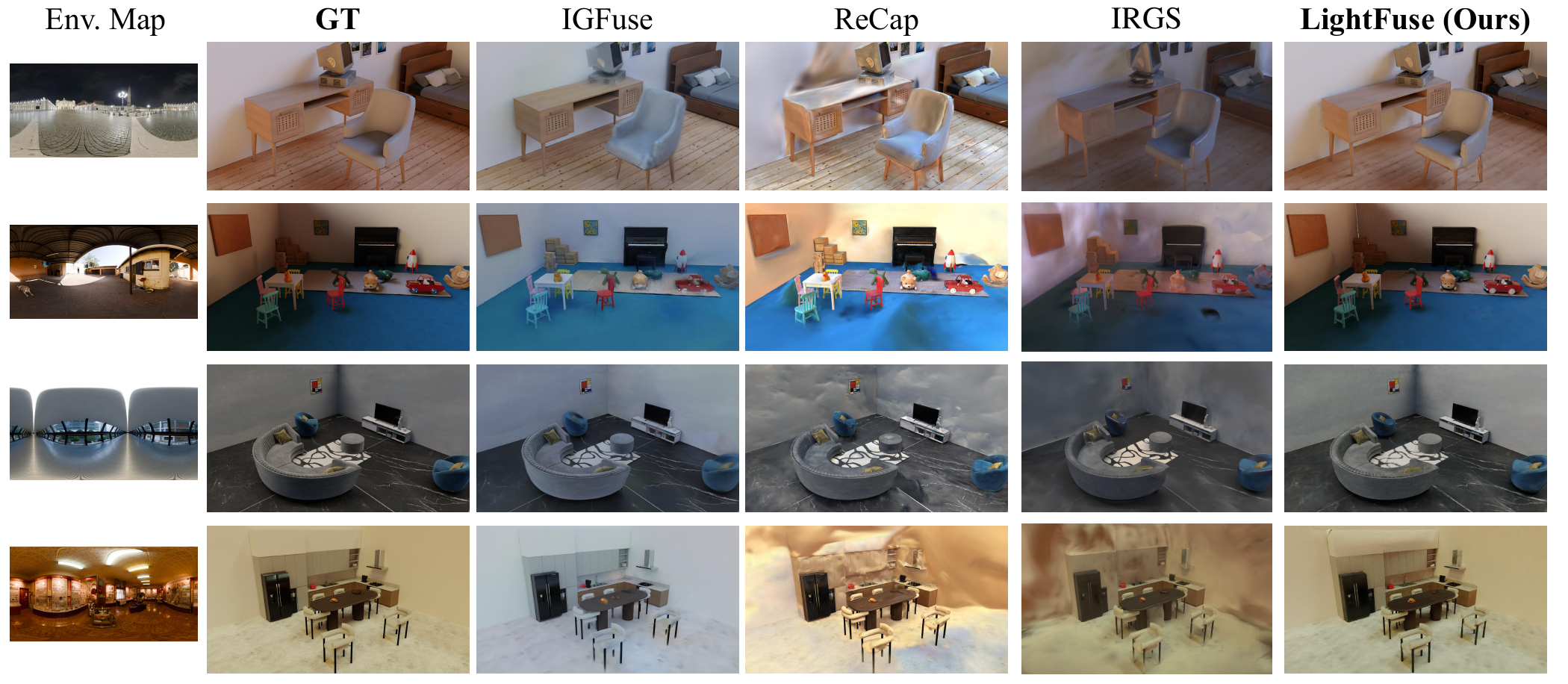}
    \caption{Novel-state relighting on Bedroom, Playroom, Livingroom, and Kitchen (top to bottom). Columns: target environment map, GT, IGFuse, ReCap, IRGS, and \method{}.}
    \label{fig:synthetic_relighting}
\end{figure*}

\begin{figure}[t]
    \centering
    \includegraphics[width=\columnwidth]{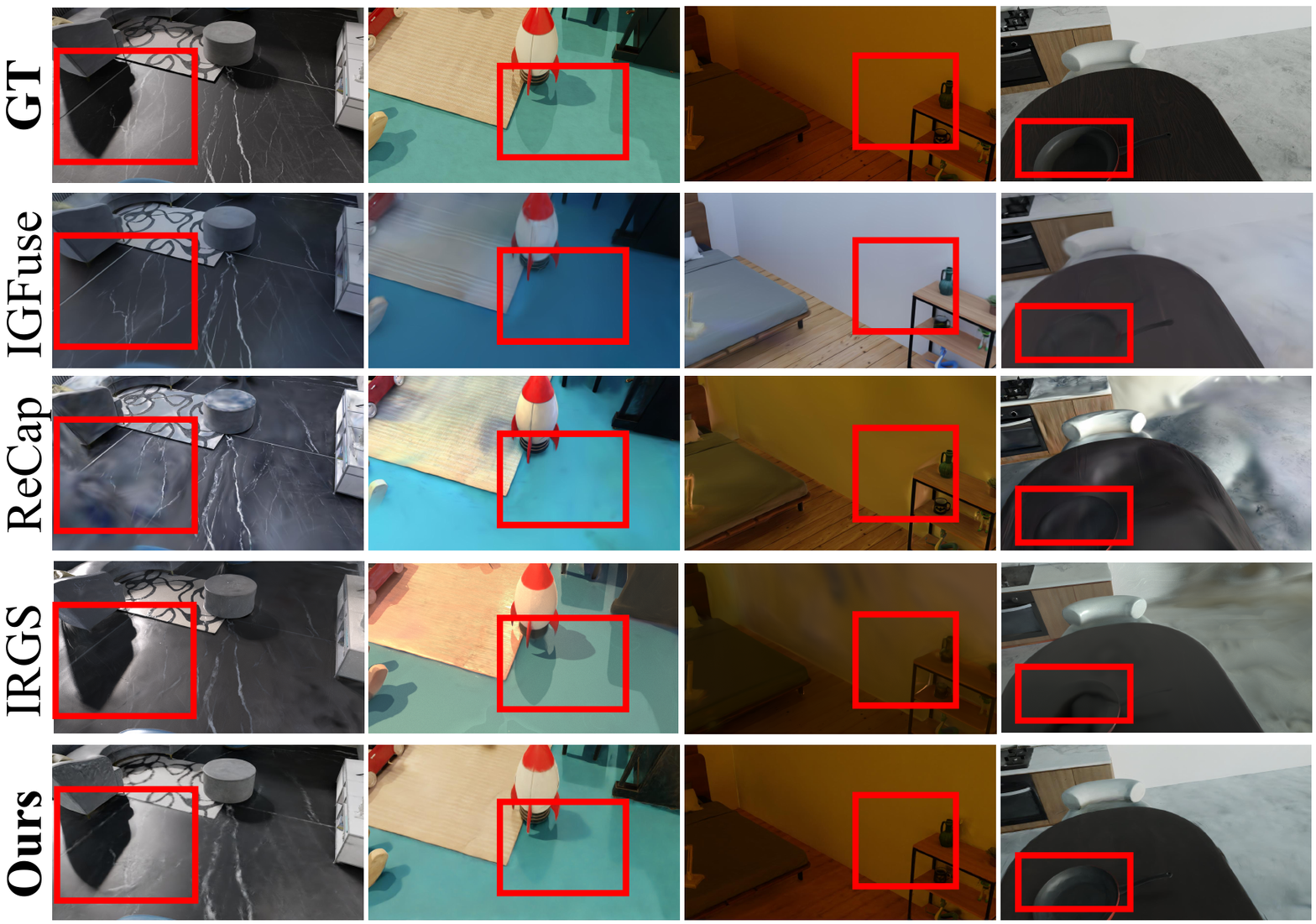}
    \caption{Detailed comparison of novel-state relighting. Red boxes highlight cast shadows, low-texture regions, and reflections.}
    \label{fig:synthetic_relighting_details}
\end{figure}

Table~\ref{tab:synthetic-comparison} compares novel-state relighting on the four synthetic scenes.
\method{} achieves 25.89\,dB PSNR and 0.873 SSIM, outperforming the strongest baseline, IRGS, by 6.01\,dB PSNR and 0.131 SSIM.
IGFuse bakes in illumination and ReCap does not trace visibility, reaching only 14.88--15.39\,dB.
Although IRGS also traces visibility, during training it queries indirect radiance from a field baked in the reference state, which is inconsistent with the layouts and illumination of the other states.
\method{} instead re-traces secondary bounces with shared materials and the state-specific $E_k$, estimates direct lighting with MIS, and fits materials jointly across all states.
Figures~\ref{fig:synthetic_relighting} and~\ref{fig:synthetic_relighting_details} show that these differences are most visible in cast shadows, low-texture surfaces, and reflections.

\begin{table}[t]
    \caption{Novel-state relighting on four synthetic scenes under eleven unseen environment maps, averaged equally across scenes. ReCap and IRGS are adapted to multi-state, multi-illumination inputs.}
    \label{tab:synthetic-comparison}
    \centering
    {\footnotesize
    \renewcommand{\arraystretch}{1.08}
    \setlength{\tabcolsep}{2pt}
    \begin{tabular*}{\columnwidth}{@{\extracolsep{\fill}}lcc@{}}
        \toprule
        Method
        & PSNR $\uparrow$
        & SSIM $\uparrow$ \\
        \midrule
        IGFuse & 14.88 & 0.714 \\
        ReCap & 15.39 & 0.713 \\
        IRGS & \underline{19.88} & \underline{0.742} \\
        \cmidrule(lr){1-3}
        \method{} & \textbf{25.89} & \textbf{0.873} \\
        \bottomrule
    \end{tabular*}
    }
\end{table}

\subsection{Novel-State Synthesis}
\label{sec:real-results}

\begin{figure*}[t]
    \centering
    \includegraphics[width=\textwidth]{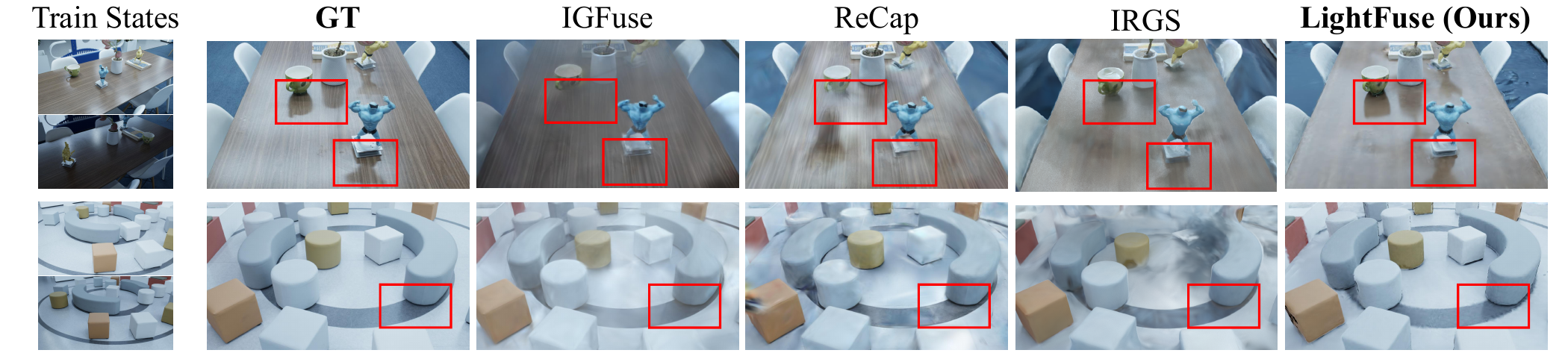}
    \caption{Qualitative comparison of novel-state synthesis on Captured 1 (top) and Captured 2 (bottom) under observed lighting. Columns: training states, GT, IGFuse, ReCap, IRGS, and \method{}. Red boxes highlight reflections and cast shadows near the moved objects.}
    \label{fig:real_synthesis}
\end{figure*}

Table~\ref{tab:real-comparison} evaluates novel-state synthesis on captured scenes under observed lighting.
\method{} achieves the highest PSNR of 24.35\,dB, outperforming the strongest baseline, ReCap, by 1.55\,dB, but has a lower SSIM (0.821 versus 0.903).
We attribute the lower SSIM to residual Monte Carlo noise and denoising blur in our path-traced renderings, to which SSIM is more sensitive than PSNR, whereas ReCap's analytic split-sum shading produces noise-free images.
Figure~\ref{fig:real_synthesis} shows \method{} better reproduces reflections and cast shadows near moved objects.

\begin{table}[t]
    \caption{Novel-state synthesis on the two captured scenes, averaged equally across scenes. ReCap and IRGS are adapted to multi-state inputs.}
    \label{tab:real-comparison}
    \centering
    {\footnotesize
    \setlength{\tabcolsep}{2pt}
    \begin{tabular*}{\columnwidth}{@{\extracolsep{\fill}}lcc@{}}
        \toprule
        Method
        & PSNR $\uparrow$ & SSIM $\uparrow$ \\
        \midrule
        IGFuse & 17.19 & \underline{0.841} \\
        ReCap & \underline{22.80} & \textbf{0.903} \\
        IRGS & 18.75 & 0.769 \\
        \midrule
        \method{} & \textbf{24.35} & 0.821 \\
        \bottomrule
    \end{tabular*}
    }
\end{table}

\subsection{Ablations}
\label{sec:ablations}

\begin{figure}[t]
    \centering
    \includegraphics[width=\columnwidth]{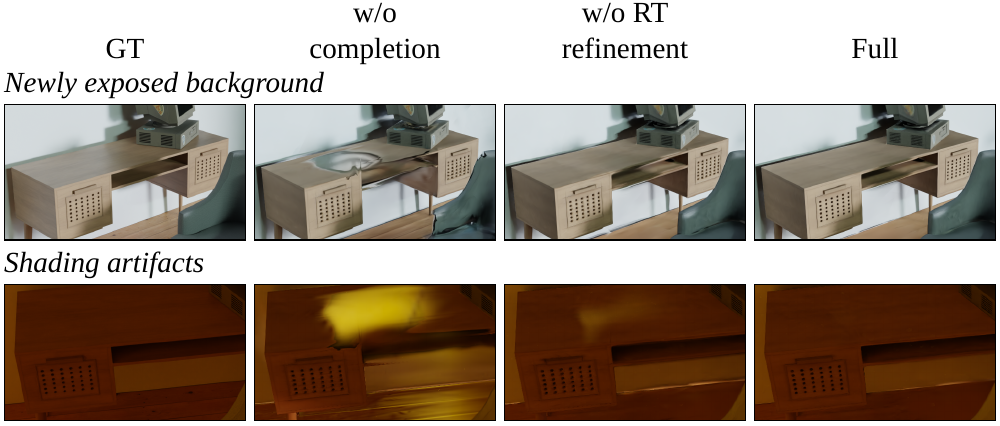}
    \caption{Effect of geometry on final relighting. Without completion, background exposed by moved objects remains missing; without ray-traced refinement, light leakage and false shadows appear.}
    \label{fig:ablation-supervision}
\end{figure}

\paragraph{Geometry pipeline.}
Ablations are averaged over the four synthetic scenes and eleven unseen environment maps but use a subset of the held-out test images, so Full results differ slightly from Table~\ref{tab:synthetic-comparison}.
Table~\ref{tab:ablation_loss} starts from reference-state geometry and cumulatively adds cross-state completion, $\mathcal L_\alpha$, $\mathcal L_s$, and ray-traced refinement, keeping depth and normal supervision and multi-state inverse rendering unchanged.
Besides relighting, we measure ray-traced ambient occlusion (AO) and normal errors against the Blender geometry, which reflect the occlusion and surfaces seen by the tracer.
Cross-state completion gives the largest AO gain (0.0434 to 0.0272 MAE) and raises PSNR by 0.93\,dB by recovering background exposed by moved objects.
$\mathcal L_\alpha$ alone degrades AO, normal error, and PSNR, but $\mathcal L_s$ then reduces normal error and raises PSNR by 1.49\,dB to 25.51\,dB.
Ray-traced refinement further lowers normal error to 1.695$^\circ$ and yields the best PSNR (25.90\,dB) and SSIM (0.875), with AO nearly unchanged (0.0304 vs.\ 0.0313).
Figure~\ref{fig:ablation-supervision} shows the resulting artifacts; individual-loss ablations are in \suppsec{sec:supp-individual-losses}.

\begin{table}[t]
    \caption{Cumulative geometry ablation. AO and normal errors are computed on the held-out layout and independent of illumination.}
    \label{tab:ablation_loss}
    \centering
    {\footnotesize
    \renewcommand{\arraystretch}{1.08}
    \setlength{\tabcolsep}{2pt}
    \begin{tabular*}{\columnwidth}{@{\extracolsep{\fill}}lcccc@{}}
        \toprule
        Geometry
        & \shortstack{AO\\MAE $\downarrow$}
        & \shortstack{Normal\\MAE ($^\circ$) $\downarrow$}
        & PSNR $\uparrow$
        & SSIM $\uparrow$ \\
        \midrule
        Reference only & 0.0434 & 4.060 & 23.28 & 0.841 \\
        + Completion & \textbf{0.0272} & 2.568 & 24.21 & 0.853 \\
        + $\mathcal L_\alpha$ & 0.0360 & 3.816 & 24.02 & 0.854 \\
        + $\mathcal L_s$ & 0.0313 & 2.580 & 25.51 & 0.870 \\
        + RT refinement (Full) & 0.0304 & \textbf{1.695} & \textbf{25.90} & \textbf{0.875} \\
        \bottomrule
    \end{tabular*}
    }
\end{table}

\paragraph{Inverse rendering.}
Table~\ref{tab:ablation_optimization} keeps the full geometry fixed and varies the inverse renderer.
Direct only removes the secondary term in Eq.~\eqref{eq:path-estimator}, and w/o MIS estimates direct lighting with NEE alone while retaining BSDF continuation for indirect transport.
Removing indirect transport lowers PSNR from 25.90\,dB to 22.54\,dB, and removing MIS lowers it to 19.86\,dB, showing both are necessary.
We compare fixed 8 and 32 spp with our progressive schedule under the same 15,000 iterations.
Fixed 32 spp attains the highest PSNR (25.97\,dB) but takes 3.01\,h, about twice the 1.59\,h of the progressive schedule, which is only 0.07\,dB lower.
Fixed 8 spp is the fastest (1.08\,h) but loses 0.24\,dB, so the progressive schedule offers the best quality--cost trade-off.

\begin{table}[t]
    \caption{Inverse-rendering ablations with fixed full geometry. Training time covers the 15,000 inverse-rendering iterations only.}
    \label{tab:ablation_optimization}
    \centering
    {\footnotesize
    \renewcommand{\arraystretch}{1.08}
    \setlength{\tabcolsep}{1.5pt}
    \begin{tabular*}{\columnwidth}{@{\extracolsep{\fill}}lccc@{}}
        \toprule
        Variant
        & PSNR $\uparrow$
        & SSIM $\uparrow$
        & \shortstack{Training\\time (h) $\downarrow$} \\
        \midrule
        Direct only & 22.54 & 0.824 & 1.35 \\
        w/o MIS (NEE direct) & 19.86 & 0.804 & 1.38 \\
        Fixed 8 spp & 25.66 & 0.874 & \textbf{1.08} \\
        Fixed 32 spp & \textbf{25.97} & \textbf{0.876} & 3.01 \\
        Full (8$\rightarrow$16$\rightarrow$32) & 25.90 & 0.875 & 1.59 \\
        \bottomrule
    \end{tabular*}
    }
\end{table}

%% file: sections/discussion_limitations.tex
\section{Limitations}

\method{} has several limitations.
Geometry decomposition relies on multi-view masks and GaussianCut, so segmentation errors corrupt object components and, with geometry fixed during optimization, propagate to material and illumination estimates; more robust segmentation could mitigate this.
Transport-aware optimization and rendering also remain expensive.
With only one indirect bounce, long-path global illumination, caustics, transmission, and mirror chains are not faithfully recovered.
Future work will target faster rendering and more expressive light transport.

%% file: sections/conclusion.tex
\section{Conclusion}

We present LightFuse, to our knowledge the first framework for relightable multi-scan interactive Gaussian scene reconstruction. It combines multi-state component reconstruction, ray-tracing-oriented geometry refinement, and inverse rendering with shared materials and state-specific illumination; retracing edited geometry supports rearrangement, material editing, and relighting with updated visibility. Experiments on synthetic and real scenes show that \method{} outperforms interactive reconstruction and inverse rendering baselines, and that ray-tracing-oriented geometry is essential for accurate relighting.

%% file: sections/llm_usage.tex
\section*{LLM Usage Statement}
Large language model (LLM)-based tools were used to assist with
code development and debugging, manuscript drafting and revision,
and experimental workflows, including experiment planning,
script preparation, and result analysis.
The authors reviewed and revised the AI-assisted code and text,
verified the reported results against experimental outputs,
and take full responsibility for the methodology, implementation,
results, and conclusions presented in this paper.